\documentclass[letterpaper, 10 pt, conference]{ieeeconf}  

\IEEEoverridecommandlockouts                              

\usepackage{graphics} 
\usepackage{epsfig} 
\usepackage{amsmath} 
\usepackage{graphicx}
\usepackage[dvipsnames]{xcolor,colortbl}
\usepackage{multirow}
\usepackage{amssymb}
\usepackage{multirow, tabularx}
\usepackage{booktabs}
\usepackage{comment}
\usepackage{hyperref}
\usepackage{subcaption}

\DeclareMathOperator{\slog}{slog}
\DeclareMathOperator{\relu}{ReLU}
\DeclareMathOperator{\sign}{sign}
\DeclareMathOperator{\mae}{MAE}
\newcommand{\etal}{\textit{et al}.}
\newcommand{\ie}{\textit{i}.\textit{e}.,}
\newcommand{\eg}{\textit{e}.\textit{g}.,}

\usepackage{pifont}
\newcommand{\cmark}{\ding{51}}%
\newcommand{\xmark}{\ding{55}}%

\title{\LARGE \bf
WasteGAN: Data Augmentation for Robotic Waste Sorting through Generative Adversarial Networks
}

\author{Alberto Bacchin$^{1 \, \dagger}$, Leonardo Barcellona$^{1}$, Matteo Terreran$^{1}$, Stefano Ghidoni$^{1}$,\\ Emanuele Menegatti$^{1}$ and Takuya Kiyokawa$^{2}$
\thanks{$\dagger$ Corresponding Author {\tt\small bacchinalb@dei.unipd.it}}%
\thanks{This work was supported by the New Energy and Industrial Technology
Development Organization (NEDO) grant number JPNP20012 and the Italian Minister for University and Research (MUR) under the initiative “PON Ricerca e Innovazione 2014 - 2020”, CUP C95F21007870007}
\thanks{$^{1}$Intelligent Autonomous System Lab, Department of Information Engineering, University of Padova, Padua, Italy}%
\thanks{$^{2}$ Department of Systems Innovation, Graduate School of Engineering Science, Osaka University, Toyonaka, Osaka, Japan.}%
}

\begin{document}

\maketitle
\thispagestyle{empty}
\pagestyle{empty}

\begin{abstract}

Robotic waste sorting poses significant challenges in both perception and manipulation, given the extreme variability of objects that should be recognized on a cluttered conveyor belt. While deep learning has proven effective in solving complex tasks, the necessity for extensive data collection and labeling limits its applicability in real-world scenarios like waste sorting. To tackle this issue, we introduce a data augmentation method based on a novel GAN architecture called wasteGAN. The proposed method allows to increase the performance of semantic segmentation models, starting from a very limited bunch of labeled examples, such as few as 100. The key innovations of wasteGAN include a novel loss function, a novel activation function, and a larger generator block. Overall, such innovations helps the network to learn from limited number of examples and synthesize data that better mirrors real-world distributions. We then leverage the higher-quality segmentation masks predicted from models trained on the wasteGAN synthetic data to compute semantic-aware grasp poses, enabling a robotic arm to effectively recognizing contaminants and separating waste in a real-world scenario. Through comprehensive evaluation encompassing dataset-based assessments and real-world experiments, our methodology demonstrated promising potential for robotic waste sorting, yielding performance gains of up to 5.8\% in picking contaminants. The project page is available at \texttt{\url{https://github.com/bach05/wasteGAN.git}}. 

\end{abstract}


\section{INTRODUCTION}
\label{sec::intro}

\begin{figure}[htbp]
    \includegraphics[width=0.99\columnwidth]{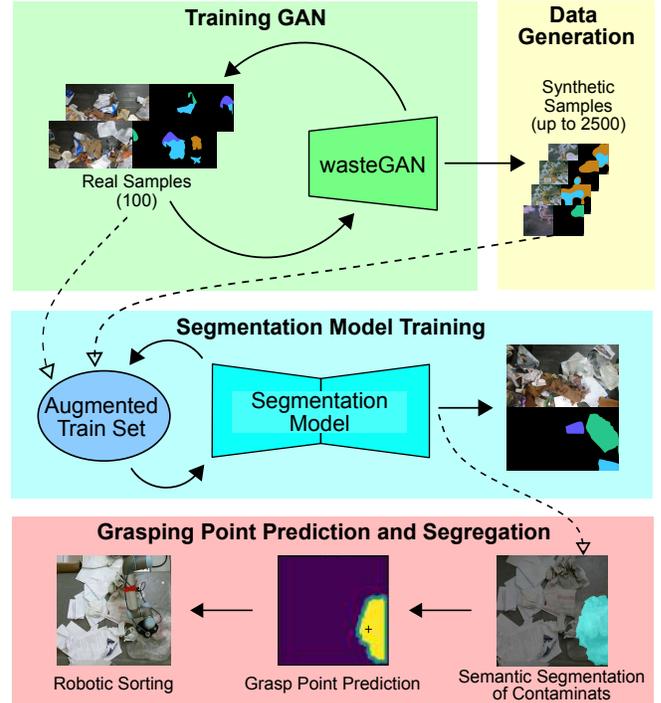}
    \caption{A general overview of the proposed pipeline for robotic waste sorting.}
    \label{fig:init}
\end{figure}

Robotic waste sorting systems aim to recognize and separate materials (such as paper, plastic, metal, and glass) in a chaotic waste stream with industrial robots~\cite{Koskinopoulou_waste_sorting}. Beyond mere automation, this technology can make an impact at multiple levels. It not only relieves humans from highly repetitive and burdensome tasks but also supports the advancement of a more circular economy. Consequently, there is a growing interest among companies~\cite{zenrobotics_evaluation} and the research community~\cite{robot_construction_sorting, pick_toss_robot_waste_sorting} in developing robotic waste sorting technologies. 

However, industrial waste sorting presents numerous challenges, including the irregular shapes of objects, high variability, and cluttered environments, all of which pose significant hurdles for perception and manipulation. Furthermore, the availability of comprehensive public datasets tailored for industrial waste management remains limited. To the best of our knowledge, ZeroWaste~\cite{zerowaste} is the only extensive dataset, collected from a paper sorting facility in the US. However, leveraging such data effectively poses significant challenges due to the inherent variations in object classes across sorting plants and the differing appearances of objects based on geographical locations~\cite{review_robotic_waste_sorting}. This challenge is particularly evident when dealing with dense annotations such as semantic segmentation masks, despite their ability to accurately represent irregular object boundaries compared to other annotation techniques like bounding boxes.

To reduce the dependency on large training sets and boost the applicability in waste sorting facilities, our work investigates the opportunity of leveraging synthetic data. Previous attempts using procedural data generation~\cite{bashkirova2023visda} proved ineffective. 
Therefore, we propose to use Generative Adversarial Networks (GANs) to generate a large amount of synthetic examples and augment the real-world data to train semantic segmentation models. GANs have been preferred due to the possibility of being trained with limited computational resources and faster compared to other generative models, such as Diffusion Models~\cite{rombach2021_diffusion_models}. Additionally, GANs have proven successful in data augmentation, although primarily in the contexts of image classification~\cite{li_eid-gan,Du_gan_surface, ma_GAN-MVA, xian_F-VAEGAN-D2} and object detection~\cite{9008794,9191127} tasks.
While there have been some efforts in applying GANs to semantic segmentation tasks, such as \textit{SemanticGAN}~\cite{li_gan_sem_seg}, these approaches have not been extensively explored in industrial settings characterized by unstructured object clusters, such as waste sorting facilities, and constrained by data scarcity. 

In response to these challenges, we introduce a novel GAN architecture called \textit{WasteGAN}. Building upon \textit{SemanticGAN}, our architecture is specifically tailored for rapid convergence with minimal labeled data in the training set. Unlike traditional GANs, our focus is not on producing photo-realistic samples, but rather on generating synthetic samples that enhance the generalization capabilities of semantic segmentation models in real-world scenarios. 
Experimental results demonstrate the effectiveness of the proposed GAN-based augmentation method when used to train different off-the-shelf semantic segmentation models on the ZeroWaste dataset~\cite{zerowaste}, the most popular benchmark for waste sorting. Compared to the baseline \textit{SemanticGAN}~\cite{li_gan_sem_seg}, our approach demonstrates superior performance across most cases.

To further showcase the effectiveness of the augmentation achieved through \textit{WasteGAN} in improving waste segregation, we integrated a semantic segmentation model trained on the augmented datasets into a real-world waste sorting system, composed of an industrial robot equipped with a vacuum gripper. Leveraging semantic segmentation of the scene, we determine optimal grasping points for target contaminant objects, enabling a robot equipped with a vacuum gripper to remove them from the waste stream. Notably, our approach addresses cluttered scenarios akin to those encountered in actual waste sorting facilities, distinguishing it from previous studies~\cite{suction_waste_robot} that usually assume well-separated objects. 
The overall scheme of the proposed approach is illustrated in Fig.~\ref{fig:init}, starting from the top you can appreciate our data augmentation technique based on \textit{WasteGAN} and the consequent grasping point generation for robotic sorting. 

In summary, our contributions are %
 (1) the introduction of \textit{WasteGAN}, a GAN architecture designed for data augmentation to address the constraints of limited annotated data and cluttered unstructured scenes inherent of the waste sorting domain; %
 (2) an extensive validation of the proposed approach by evaluating different state-of-the-art segmentation models trained on \textit{wasteGAN} output, demonstrating an overall improvement in semantic segmentation performance when compared with other GANs synthetic outputs; %
 (3) integration of the proposed augmentation model in a real robotic waste sorting system, exploiting the segmentation mask to extract a suitable grasping point for an industrial manipulator; %
 (4) evaluation of the entire proposed methodology in a real paper sorting application developed in a mockup environment in our laboratory, highlighting the robustness and generalization capabilities of the model trained with \textit{wasteGAN} when used in a different setting than the one of the training data.

\section{RELATED WORKS}
\label{sec:related_works}

\subsection{Robotic Waste Sorting Systems}

A Robotic Waste Sorting System (RoWSS) is composed of two key elements: (1) a perception system tasked with identifying various types of waste within the working area and (2) a robotic arm to physically pick and segregate waste. RoWSS can be broadly categorized into (a) mobile platforms~\cite{robot_construction_sorting} able to collect waste from the environment and (b) industrial sorters intended for installation in waste management facilities~\cite{zenrobotics_evaluation}.

Our focus in this study is primarily on the latter category. Koskinopoulou~\etal~\cite{robotic_waste_sorting_system} acknowledges the challenge posed by data scarcity and the difficulty of generalizing to new scenarios. The proposed solution involves a data augmentation technique that randomly pastes patches of waste images onto random backgrounds. Despite the promising results, their testing scenario lacks cluttered scenes and the variance of objects remains limited to those present in the original dataset. Conversely, Kiyokawa~\etal~\cite{robotic_sorter_kiyokawa} design a semi-automatic data collection procedure, but again applicable for single object detection. Using a generative model, we want to go beyond these limitations, dealing with cluttered scenarios and increasing the variance of the original dataset. 
In addition to perception, manipulating waste poses a significant challenge due to the large variability in the shape and dimension of objects. Ku~\etal~\cite{construction_demolotion_waste} propose to use depth images from an RGB-D camera to sample grasp poses, represented as grasp rectangles~\cite{grasp_rectangle}, and score them with a neural network to extract the most suitable one. However, this method is effective only when the target object is clearly distinguishable from the background in the depth image, thus impractical for highly cluttered scenarios. Um~\etal~\cite{suction_waste_robot} instead focuses on suction gripper technology, introducing a scoring algorithm to evaluate the quality of grasping poses on irregular surfaces. Despite the good results, this method passes over the semantics of the scene and solely focuses on computing the most suitable grasping pose. Recognizing the importance of waste classification in the sorting process alongside grasp success rate, we propose leveraging semantic segmentation masks to compute suitable grasping poses in our work.

\subsection{Data Augmentation with GAN}

Generative Adversarial Networks (GAN) have been introduced by Goodfellow~\etal~\cite{gan_bengio}. The proposed model was meant to generate new samples that were indistinguishable from a set of training examples. Many milestone improvements have been developed, such as \textit{Conditional GAN} (C-GAN)~\cite{mirza2014_cgan}, \textit{Wasserstein GAN}~\cite{wgan-arjovsky17a}, \textit{Progressive-Growing GAN}~\cite{karras2018_pggan} and \textit{StyleGAN}~\cite{style_gan,stylegan2}. These models have demonstrated the capability to generate good-quality images in various scenarios, inspiring data augmentation techniques based on GANs. Most of the works are focused on classification because applying GANs in this domain is straightforward: it is possible to train multiple networks on different classes to generate per-class synthetic samples and train a classifier~\cite{Bissoto_2021_CVPR}. 
However, this approach is very time-consuming when the number of classes increases. 
\textit{Conditional GAN} solves the issue by using the label as a prior to synthesize images of a specific class, allowing to generate a labelled classification dataset~\cite{covid_Waheed,fruit_BIRD}.
When it comes to more complex tasks, such as semantic segmentation, GANs have been applied in different ways. In the agriculture scenario, Fawakherji~\etal~\cite{crop_Fawakherji} used segmentation labels to condition a C-GAN in order to generate new samples of crops of the rare classes in the dataset to reduce the class imbalance. However, this strategy is not suitable in the waste sorting scenario where many objects of different shapes and classes are cluttered. 
Li~\etal~\cite{li_gan_sem_seg} proposed a \textit{StyleGAN}-based architecture, named \textit{semanticGAN}, to achieve weak supervised semantic segmentation. Since images and labels are strictly correlated, the authors wanted to exploit the knowledge accumulated by the GAN trained on the unlabelled dataset to ease the generation of labels for an unknown query image. To do that, authors performed GAN inversion~\cite{xia_gan_inversion} to map the query image into the latent space of \textit{StyleGAN} and then they drove the network to reconstruct the original image alongside a coherent segmentation mask. 
While this strategy has been demonstrated effective, it is slow in inferring segmentation masks, reducing its applicability in waste sorting scenarios where real-time processing is required.
To overcome these limitations, in this work we investigate several architectural changes to \textit{SemanticGAN}, focusing on generating images and segmentation masks to be used as training data for semantic segmentation models. A similar approach was explored in \textit{DatasetGAN}~\cite{datasetgan_zhang}, where the latent code of a \textit{StyleGAN}-based architecture was used to generate semantic segmentation labels using a lightweight ensemble of classifiers. However, this idea relies on the assumption that the latent code reflects certain structures in the image (e.g., the position of eyes in a face) and fails in unstructured scenarios, such as randomized clusters of waste.  

\section{METHOD}
\label{sec:method}

Our hypothesis posits that by augmenting a small set of labeled data (\eg~100) with generated samples, we can alleviate the burden of data collection while simultaneously improving model performance. In pursuit of this objective, the proposed \textit{wasteGAN} introduces architectural innovations, including an optimized activation function, an enhanced generator, and a novel loss function. Leveraging semantic segmentation masks, we compute grasping poses to effectively separate waste with an industrial robot. 

\subsection{GAN Architecture}

\begin{figure*}[tb!]
\centering
\resizebox{0.95\linewidth}{!}{
\includegraphics[]{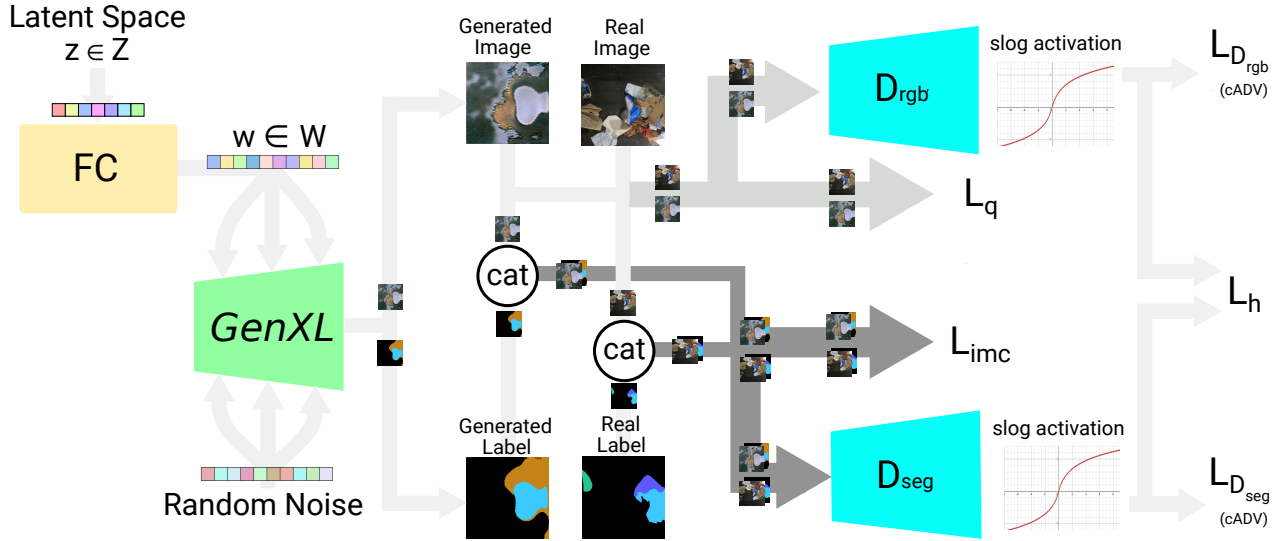}
}
\caption{The overall structure of the proposed \textit{wasteGAN}. In particular, we highlight the usage of the newly proposed methods, \ie the improved generator (\textit{GenXL}), the \textit{image-label correlation loss} (\textit{$\mathcal{L}_{imc}$}), the custom adversarial loss (\textit{cADV}) and the \textit{quality loss} (\textit{$\mathcal{L}_q$}).}
\label{fig:model}
\end{figure*}

We designed a GAN architecture, evolved from previous work~\cite{li_gan_sem_seg, stylegan2}, to effectively leverage limited amount of real-world data. The architecture of the proposed \textit{wasteGAN} is shown on Fig.~\ref{fig:model}. Starting from \textit{SemanticGAN}~\cite{li_gan_sem_seg}, we refine the architecture of the generator $G(w) : \mathcal{W} \rightarrow \mathcal{X_G} \times \mathcal{Y_G}$, where $w$ is a latent variable and $X_G$ , $Y_G$ are the synthetic image-label couples.
Since we do not need to reconstruct the original image, we removed the unnecessary GAN-inversion encoder while retaining the fully connected mapping network $FC : \mathcal{Z} \rightarrow \mathcal{W}$ introduced in~\cite{stylegan2}. Instead, we enhanced the generator's capabilities by augmenting the size of its blocks, achieved by stacking an additional style layer atop the original two. This refined generator is denoted as \textit{GenXL}.

We made use of two discriminators as in \textit{SemanticGAN}~\cite{li_gan_sem_seg}: (1) $D_{rgb} : \mathcal{X_R} \cup \mathcal{X_G} \rightarrow  \mathbb{R} $ takes real ($\mathcal{X_R} $) or generated ($\mathcal{X_G}$) images and gives a score to discriminate between them; (2) $D_{seg} : (\mathcal{X_R} , \mathcal{Y_R}) \, \cup \, (\mathcal{X_G} ,  \mathcal{Y_G}) \rightarrow  \mathbb{R} $ takes the concatenation of images and labels. While $D_{rgb}$ favors the generation of high quality images, $D_{seg}$ encourages the generation of consistent image-label pairs. We used a residual architecture for both discriminators, as in~\cite{stylegan2}.

Originally, the last layer was a simple linear combination of previous layer's activations. Since discriminators are binary classifiers, the higher the confidence in the prediction, the larger the output score. Linear activation implies a quicker divergence and the overfitting of the discriminator, especially when the training set is small. To reduce this effect it is possible to use bounded activation like $\tanh()$, which however suffers of the zero-gradient problem. For these reasons, we designed the \textit{Symmetric Logarithm} function, defined as:
\begin{equation}
  \slog (x) = \begin{cases}
  \log(ax+1)  & x \ge 0 \\
  -\log(-ax+1) & x < 0
\end{cases},
\end{equation}
where $a>0$ is a hyperparameter. The $\slog()$ function is differentiable in $\mathbb{R}$, and it better controls divergence in case of large input activation since it follows a logarithm trend, while mitigating the zero-gradient problem because the gradient never goes to zero. 

\subsection{Loss Function}

The \textit{WasteGAN} adversarial loss comes from the hinge loss~\cite{lim2017_geometric_gan, chao_contrained_GAN}, we refer to this as \textit{cADV}. For the discriminators, we introduce the following hinge losses: 
\begin{multline}
    \mathcal{L}_{D_{rgb}} = \underset{x \in \mathcal{X_R}}{\mathbb{E}} \Bigl[ \relu \bigl( k-D_{rgb}(x) \bigr) \Bigr] + \\
    \underset{{x} \in \mathcal{X_G}}{\mathbb{E}}  \Bigl[ \relu \bigl( k+D_{rgb}({x})\bigr) \Bigr],
    \label{eq:drgb_loss}
\end{multline}
and
\begin{multline}
    \mathcal{L}_{D_{seg}} = \underset{x \in \mathcal{X_R} \cup \mathcal{Y_R}}{\mathbb{E}} \Bigl[ \relu \bigl( k-D_{seg}(x) \bigr) \Bigr] + \\
    \underset{{x} \in \mathcal{X_G} \cup \mathcal{Y_G}}{\mathbb{E}}  \Bigl[ \relu \bigl( k+D_{seg}({x})\bigr) \Bigr], 
    \label{eq:seg_loss}
\end{multline}
with $k \in [0,1]$. The optimal value of $\mathcal{L}$ is $D^*(x) = \sign(P_R(x) - P_G(x)) = \pm k$. Usually $k$ is set to 1, such that the discriminator tends to converge to the  $D^*(x) = \pm1$. We instead set $k$ to 0.5 to shift the equilibrium point towards  $D^*(x) = \pm0.5$ to stay further from the low-gradient area. 

For the generator, we used a loss composed of 3 terms:
\begin{equation}
    \mathcal{L}_G = \mathcal{L}_h + \mathcal{L}_q + \mathcal{L}_{imc},
    \label{eq:gen_loss}
\end{equation}
where $\mathcal{L}_h$ is an hinge adversarial loss, $\mathcal{L}_q$ is the \textit{quality loss} and $\mathcal{L}_{imc}$ is the \textit{image-label correlation loss}. The first term in Eq~\ref{eq:gen_loss} is an hinge loss, taking in account of both discriminators.

\begin{multline}
        \mathcal{L}_{h} = - \alpha  \underset{{x} \in \mathcal{X_G}}{\mathbb{E}}  \Bigl[ D_{rgb}({x})\Bigr]  + (1-\alpha) \underset{x \in \mathcal{X_R}}{\mathbb{E}} \Bigl[  D_{rgb}(x)  \Bigr] + \\
        - \alpha  \underset{{x} \in \mathcal{X_G} \cup \mathcal{Y_G}}{\mathbb{E}}  \Bigl[ D_{seg}({x})\Bigr]  + (1-\alpha) \underset{x \in \mathcal{X_R} \cup \mathcal{Y_R} }{\mathbb{E}} \Bigl[  D_{seg}(x)  \Bigr].
\label{eq:gh_loss}
\end{multline}

The hinge loss $\mathcal{L}_h$, expanded in Eq~\ref{eq:gh_loss}, penalizes the generator when the discriminator correctly classify an image, whether it is real or synthetic. we employed an unbalanced hinge loss to better manage overfitting on the real data. We empirically set $\alpha=0.8$, placing more emphasis on the first term to achieve a better balance in training.

The second loss term in Eq.~\ref{eq:gen_loss}, $\mathcal{L}_q$ or \textit{quality loss}, is used to encourage high quality on synthetic images and is composed of two terms: $\mathcal{L}_q = \mathcal{L}_p + \mathcal{L}_s $. The first term $\mathcal{L}_p$ is a perceptual loss~\cite{perceptual_loss} based on VGG features. Since GAN tends to generate image with smooth edges, we also added $\mathcal{L}_s = \mae \bigr( \, \mathfrak{S}(x_R) - \mathfrak{S}(x_G) \, \bigl)$, where $\mae (\cdot)$ is the mean absolute error and $\mathfrak{S}(x) = \sqrt{ (x \circledast S_x )^2 + (x \circledast S_y )^2 }$ is a measure of the sharpness through Sobel operator. 

Finally, $\mathcal{L}_{imc}$ or \textit{image-label correlation} loss has been designed to promote the correlation between synthetic images and labels. It is defined as a mean absolute error as follows:

\begin{equation}
    \mathcal{L}_{imc} = \mae \bigr( \, P_R(p_{i} \, | \, l_j) - P_G({p}_{i} \, | \, {l}_j) \, \bigl),
\end{equation}
where the conditional distributions describes the probability that a pixel with value $p_i$ is assigned to the label $l_j$, respectively in the real and generated image-mask couples.

By minimizing $\mathcal{L}_{imc}$ loss, we aim to maintain the relationship between labels and masks we have in real samples also in the generated samples. Since $P_G({p}_{i} \, | \, {l}_j)$ cannot be computed a priori, we compute on the fly the probability distributions using the image-mask couples in each mini-batch and then we updated the actual distribution estimations with exponential moving average (EMA). This term in the loss function is also meant to retain annotations for the infrequent classes, balancing the convergence to produce more varied labels.

\begin{figure*}[tb!]
\centering
\resizebox{0.95\linewidth}{!}{
    \includegraphics{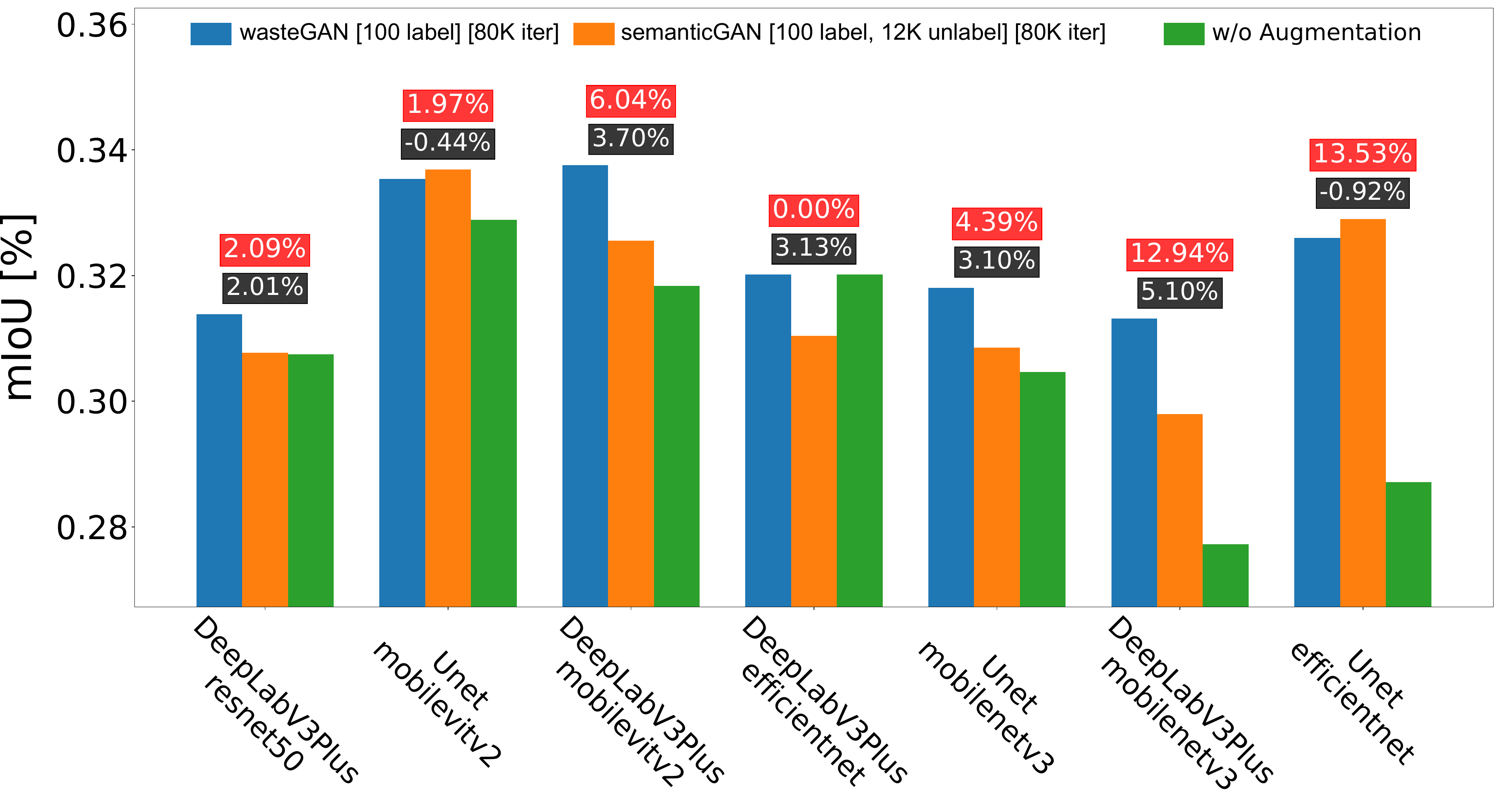}
    }
    \caption{Results from testing the models trained on synthetic datasets generated with \textit{semanticGAN}~\cite{li_gan_sem_seg} (orange), our approach (blue) and the dataset with only 100 real samples. Black values are the performance gains of \textit{wasteGAN} with respect to \textit{semanticGAN}. Red values are the performance gains with respect to training on 100 samples.}
    \label{fig:test_143k}
\end{figure*}

\subsection{Training procedure}

Our model has been implemented in \textit{PyTorch}, starting from the implementation of \textit{StyleGAN2}~\cite{stylegan2} from~\cite{labml}. We trained our GAN model with ADAM optimizer with learning rate $10^{-4}$ for the discriminators and the generator and $10^{-6}$ for the mapping network $W$. At each training iteration, we update discriminators $D_{rgb}$ and $D_{seg}$ first and generator $G$ afterwards. Since discriminators are updated using the weights of $G$, in Eq.~\ref{eq:gh_loss} the gradient of second and forth terms is non-zero.

We trained our model for 80K steps on a single NVIDIA RTX 4090 GPU. We did not apply gradient penalty~\cite{gradient_panality} since the architectural changes proposed in Sec.~\ref{sec:method} can control gradients, avoiding the slow down in training time. It is worth noticing that in the previous work of Li~\etal~\cite{li_gan_sem_seg}, the gradient propagation from $D_{seg}$ to $G$ was stopped since the authors aimed to align the generated label with the synthesized image, as opposed to altering the image creation process to fit the labels, in order to generate a correct label for the input image. In our case, however, we do not need this constraint since we want to generate both images and labels.

We trained the proposed \textit{wasteGAN} architecture on a dataset of 100 labeled examples taken from ZeroWaste~\cite{zerowaste}. The dataset contains images from a real-world paper sorting plant and manually labelled with semantic segmentation masks of 5 classes (background+paper, rigid plastic, cardboard, metal, soft plastic). For comparison, we trained \textit{semanticGAN}~\cite{li_gan_sem_seg} too. Since it requires a weakly supervised setting, we prepared an additional unlabelled set of data of 11780 images obtained gathering the rest of ZeroWaste~\cite{zerowaste} and part of ZeroWaste-v2~\cite{bashkirova2023visda} dataset. The latter is acquired in a paper sorting plant as well, but it does not provide any label. 

\subsection{Grasp Point Inference}
\label{sec:grasp_point_inference}
The generator $G$ produces synthetic samples ${(x_G, y_G) \in X_G \cup Y_G }$, which are combined with a limited amount of real-world data to train state-of-the-art real-time semantic segmentation networks to obtain better performance. The segmentation masks derived from this process are subsequently employed to determine suitable grasping points for a robot equipped with a suction cup, enabling the effective segregation of contaminants.

To compute grasping points, we considered two requirements: (1) the grasping point has to belong to the contaminant with high confidence and (2) the grasping point should be near the center of the object to facilitate a stable grip. To fulfill these requirements, we design a simple yet effective algorithm. To optimize the positioning of the suction cup within each segmented object, we apply erosion to the boundaries of the segmentation mask using a kernel size matching the radius of the suction cup. This ensures that the cup is accurately placed within the object boundaries. Next, to determine the centroid of irregularly shaped objects, we iteratively applied an erosion morphological operator until all clusters in the mask were reduced to single pixels. We associate a score to each point based on the logits predicted by the network. The final grasping point prediction will be the point with the highest score \ie~the point with the higher confidence in category prediction. Some examples are shown in Fig.~\ref{fig:real_world_images}. Note that it is also possible to consider the top-k points with the highest scores to compute grasping points for k objects.

\section{Experiments}
\label{sec:experiments}

To evaluate the effectiveness of \textit{WasteGAN} for data augmentation from limited annotated samples, we articulated experiments both on ZeroWaste~\cite{zerowaste} dataset and in real-world. 

\subsection{Evaluation on Dataset}

The evaluation pipeline on dataset is composed of four phases: (A) the training of a GAN model on the small set of 100 random examples from ZeroWaste~\cite{zerowaste} training set; (B) the creation of a synthetic dataset by random sampling from the GAN generator $G$; (C) the training of several off-the-shelf semantic segmentation models~\cite{Iakubovskii_smp} with different augmentation ratios (\ie~the ratio between the synthetic and real examples used in training process); (D) the testing of the trained models on a set of unseen examples. For the latter, we employed the test set of \textit{ZeroWaste}~\cite{zerowaste} and the mIoU as a metric. 
For testing, we considered 8 different segmentation models combining the popular Unet~\cite{fpn} and DeepLabV3+~\cite{deeplabv3plus2018} heads with Resnet50~\cite{resnet}, EfficientNet~\cite{tan2019efficientnet}, MobileNetV3~\cite{mobilenetv3} and MobileViTV2~\cite{mobilevitv2} backbones, pretrained on ImageNet-1k~\cite{imagenet}. 

Following the aforementioned evaluation procedure, we trained the semantic segmentation models, with augmentation rates ranging from $0$ (\ie~only 100 real examples) to $25$ (\ie~2500 synthetic examples in addiction to the 100 real samples already available). 
We trained \textit{semanticGAN}~\cite{li_gan_sem_seg} on \textit{ZeroWaste}~\cite{zerowaste} for comparison. The~\textit{semanticGAN} architecture is one of the few works which rely on GANs to improve semantic segmentation performance under our same hypothesis of having only a few labelled data. Nevertheless, authors in~\cite{li_gan_sem_seg} made use of an additional set of unlabelled samples which our \textit{wasteGAN} does not require. After training \textit{semanticGAN} with the additional unlabelled data, we followed the aforementioned pipeline from point (B). 

Once completed, we collected the results in Fig.~\ref{fig:test_143k}. Despite that the effect of the augmentation depends on the underlying architecture, we can observe that in most of the cases our \textit{wasteGAN} performs better with an average gain of 2.2\% with respect to \textit{semanticGAN} and a 5.8\% improvement against training on real samples only. 

\begin{figure}[tb!]
    \includegraphics[width=0.99\columnwidth]{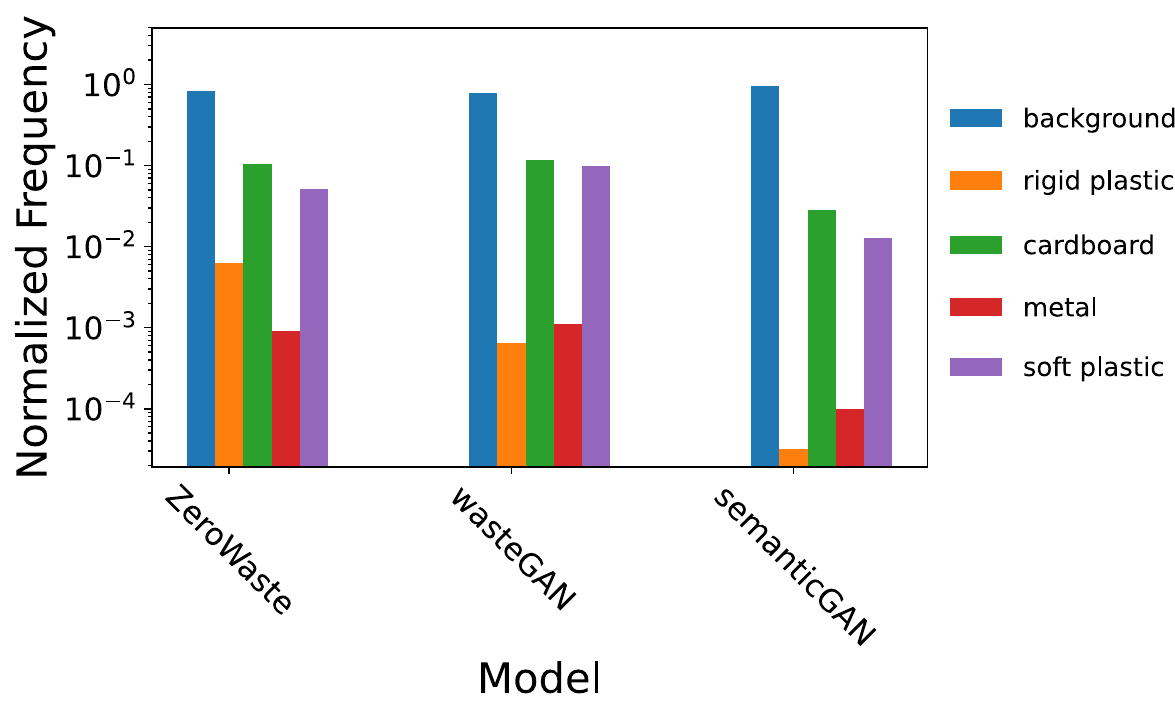}
    \caption{Frequencies of the labels in the real-world dataset (ZeroWaste~\cite{zerowaste}) and the generated datasets with \textit{semanticGAN}~\cite{li_gan_sem_seg} and the proposed \textit{wasteGAN}. Scale is logarithmic.}
    \label{fig:class_distributions_143k}
\end{figure}

Evaluation on different semantic segmentation models showed the generalization capability of our approach while providing better results than state-of-the-art GAN-based solutions. We tried to investigate the reason behind this success by analyzing the label distribution in the real and synthetic datasets. In Fig.~\ref{fig:class_distributions_143k} we depict the label frequencies of the full training set of ZeroWaste~\cite{zerowaste} and of the datasets generated with \textit{wasteGAN} and  \textit{semanticGAN}. It is easy to see that the proposed method can better reproduce the original distribution, even for less frequent classes of contaminants like ``metal'' or ``rigid plastic''. 

The precise label distribution in \textit{wasteGAN} is a key factor contributing to the improvements demonstrated by models trained on \textit{wasteGAN} synthetic images. This advantage over \textit{semanticGAN} primarily stems from the two modifications implemented in the \textit{wasteGAN} generator. Firstly, the inclusion of an additional style block in the \textit{GenXL} generator enhances the expressiveness of synthetic images. Secondly, the introduction of quality loss and image-label correlation loss terms in the generator's loss function enables a more accurate capture of the label distribution, especially for underrepresented classes.

Finally, it is worth to highlight that the absence of the encoder to perform GAN inversion in \textit{wasteGAN} brings to a significant 20x speed-up in generation time, from $2.2 s$ to $0.12 s$, reducing the time needed to create new synthetic data and boosting the applicability in the real world scenarios.

\begin{table}[tb!]
  \centering
  \begin{tabular}{cccc}
    \toprule
    Aug Mode & $A_C$ & $A_G$ & $FPR$ \\
    \midrule
    \midrule
     \textit{No Aug} & 0.44 & 0.29 & 0.3 \\
     \textit{semanticGAN}~\cite{li_gan_sem_seg} & 0.32 & 0.18 & 0.7 \\ 
    \midrule
     \textit{wasteGAN} (ours) & \textbf{0.45} & \textbf{0.35} & \textbf{0.2} \\
    \bottomrule
  \end{tabular}
  \caption{Comparison of the impact of different augmentation modalities on a real-world pick and place task. Bold indicates best scores. \textit{No Aug} indicates a model trained on 100 real world samples. While \textit{semanticGAN} and \textit{wasteGAN} indicates that the model has been trained also with synthetic data. }
  \label{tab:real_worl_results}
\end{table}

\begin{figure}[tb!]
    \includegraphics[width=0.99\columnwidth]{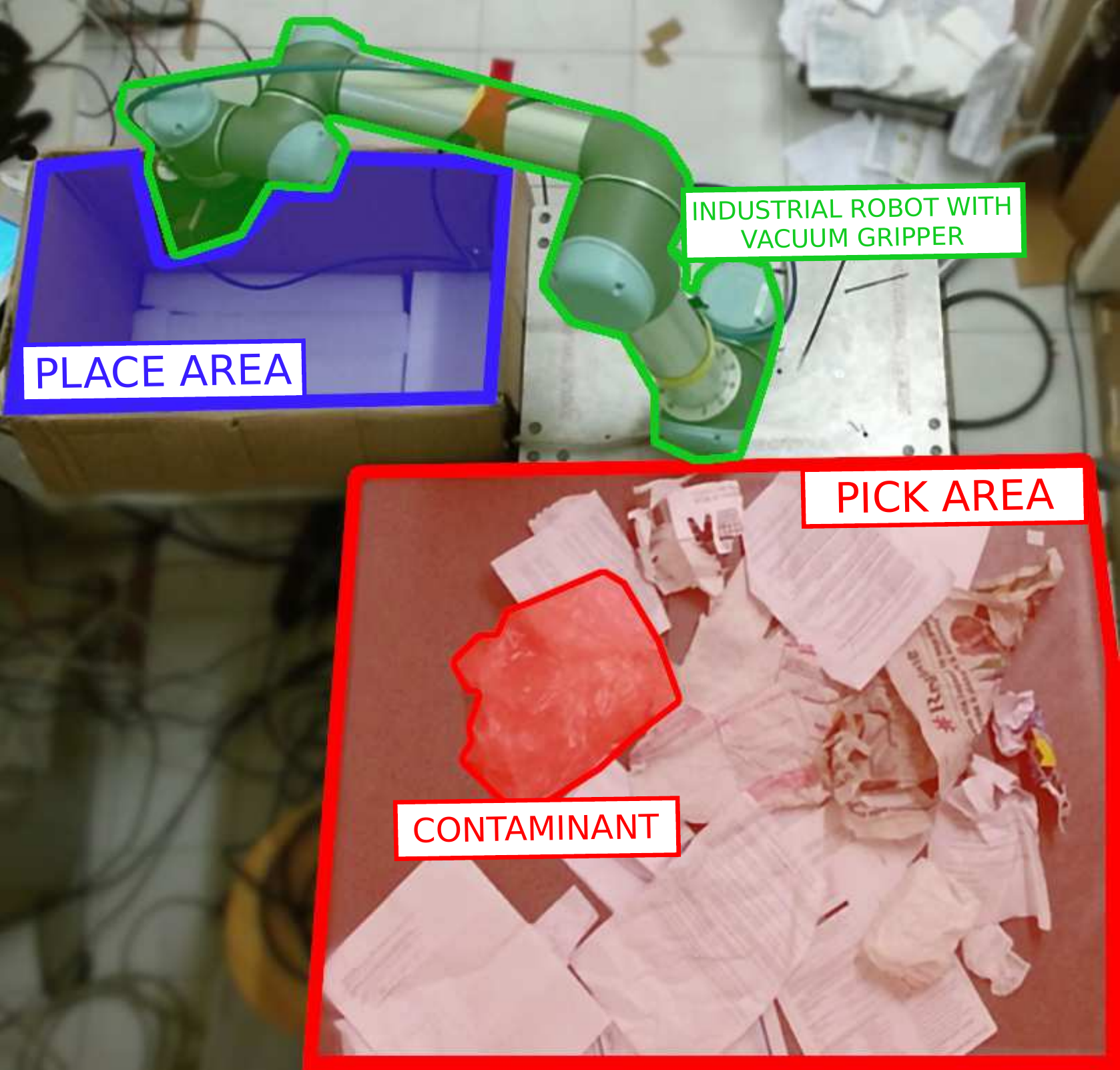}
    \caption{Our experimental setup.}
    \label{fig:setup}
\end{figure}

\subsection{Evaluation with a Robotic Waste Sorting System}

The experiments with the real robot had the objective to validate the grasping point prediction, in particular evaluating the impact of the GAN-based augmentation pipeline on the success rate of a pick and place task. The experiments have been held using a custom 3D-printed vacuum gripper applied to a UR5 robot and a KinectV2 RGB-D camera. We used the model showing best performances evaluated on ZeroWaste~\cite{zerowaste} \ie~a DeepLabV3Plus + MobileViTV2 model.

After predicting the grasping point in the image space, we use the depth map to project the point into the robot space and execute the motion. Mirroring the composition of ZeroWaste~\cite{zerowaste}, we collected real samples of waste belonging to the same five categories of the dataset. The sorting task consists in removing contaminants from paper waste in a cluttered scenario and drop into the place area, see Fig.~\ref{fig:setup}.

For each run, we randomly place some paper waste as a background in the robot's workspace and we placed a contaminant of specific category inside the clutter. To fairly compare the 3 different augmentation modalities, we replace the target object in the same position without changing the background. To evaluate each run we take in account of (1) the accuracy in recognizing the contaminant and its category through the semantic segmentation model ($A_{C}$) and (2) the accuracy in picking it and drop it in the place bin ($A_G$). We performed a total of $58~\times~3$ runs with different contaminants. Additionally, we performed $10 \times 3$ runs with solely paper waste in the working space to evaluate the false positive rate of the models ($FPR$).

\begin{figure}[tb!]
    \includegraphics[width=0.99\columnwidth]{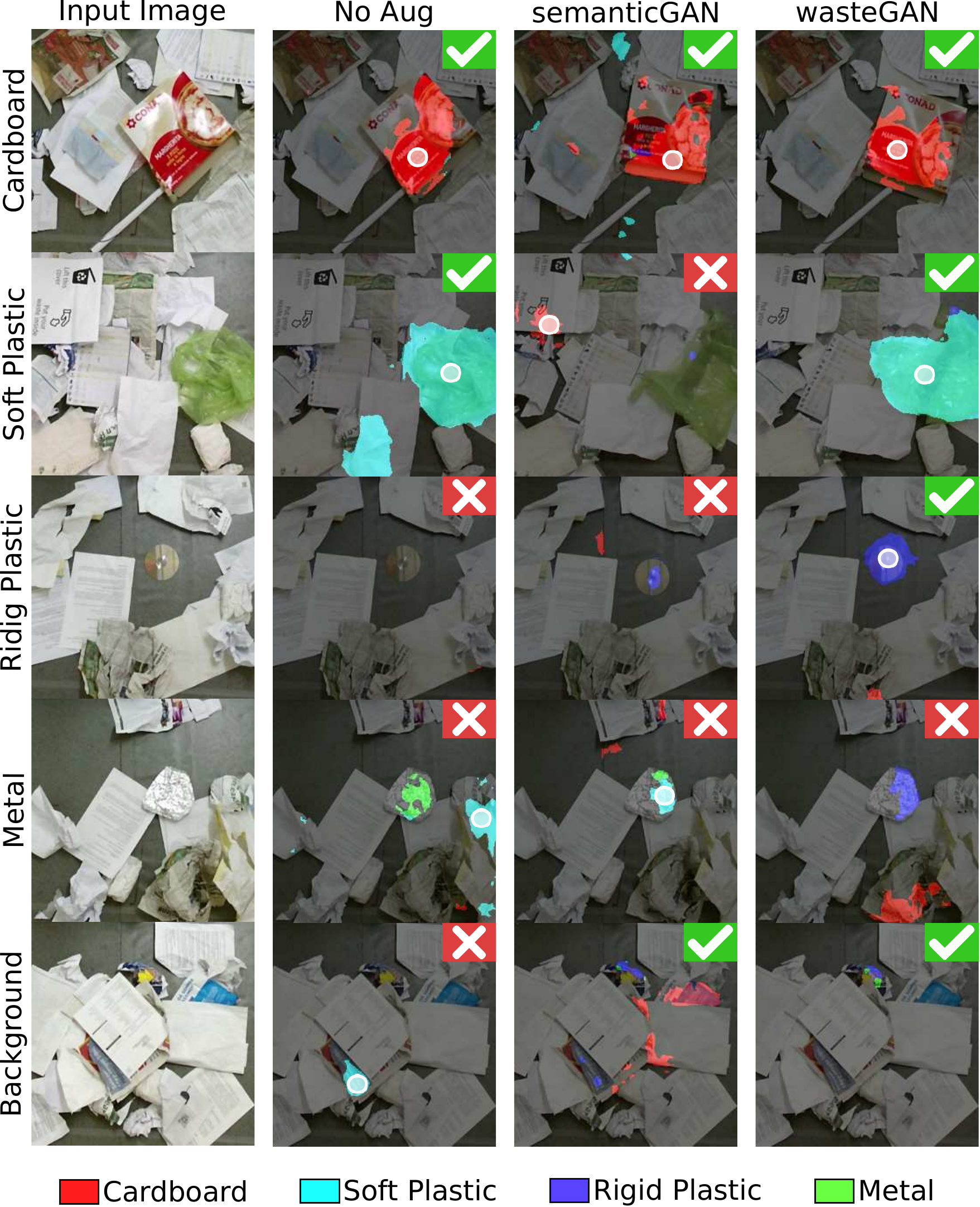}
    \caption{Examples images taken from real world experiments. Each row represents a target contaminant or background. The white circle represents the predict grasp point (if any). Colors represent the segmented areas for each class. You can appreciate both success (~\cmark~) and failure (~\xmark~) cases. }
    \label{fig:real_world_images}
\end{figure}

Results are shown in Table~\ref{tab:real_worl_results}, alongside prediction examples in Fig.~\ref{fig:real_world_images}. It can be appreciated how the proposed augmentation method significantly enhances both recognition and grasping accuracy. While the quality of the images generated by \textit{semanticGAN} seems to penalize the performance of the model, our \textit{wasteGAN} is able to increase the generalization capabilities of the segmentator on a different scenario. Although the performance in recognizing the target object are comparable between \textit{wasteGAN}-augmented and not augmented model, the higher mask quality provided by the former lead to a better grasping point prediction and so an higher accuracy for the whole pick and place task. Notably, the augmentation process with the proposed pipeline is also able to reduce the $FPR$ which is important for real-world application in order to avoid useless robot working cycles.  

\section{Conclusions}
\label{sec:conclusion}

This work presents an innovative approach for robotic waste sorting, addressing challenges of data scarcity and cluttered scenes. Our focus lies in optimizing a GAN architecture to facilitate training with limited data while preserving real distribution characteristics. By leveraging augmented datasets with synthetic samples, we train a semantic segmentation model and utilize the generated masks to compute semantic-aware grasping poses. This enables the deployment of a real robotic waste sorter, allowing us to evaluate both the model's generalization capabilities in real-world scenarios and the performance of the proposed semantic-aware grasping pose computation. The results confirm that the data augmentation process through \textit{wasteGAN} significantly improve the performance in the real-world experiments, highlighting the capabilities of the proposed approach to tackle the domain shift between different settings.  

Despite the promising results, we acknowledge the potential for further enhancements. To improve the effectiveness of data augmentation, our plan is to shift the augmentation process from the image space to the feature space. Working in a latent space has demonstrated a successful strategy with others generative models~\cite{rombach2021_diffusion_models}. Additionally, we intend to explore advanced grasping point computation techniques, for example integrating our semantic-aware approach with existing class-agnostic methods~\cite{suction_waste_robot} and better leveraging the degrees of freedom of a robotic arm.










\bibliographystyle{IEEEtran}
\bibliography{bibliography.bib}

\begin{thebibliography}{10}
\providecommand{\url}[1]{#1}
\csname url@samestyle\endcsname
\providecommand{\newblock}{\relax}
\providecommand{\bibinfo}[2]{#2}
\providecommand{\BIBentrySTDinterwordspacing}{\spaceskip=0pt\relax}
\providecommand{\BIBentryALTinterwordstretchfactor}{4}
\providecommand{\BIBentryALTinterwordspacing}{\spaceskip=\fontdimen2\font plus
\BIBentryALTinterwordstretchfactor\fontdimen3\font minus \fontdimen4\font\relax}
\providecommand{\BIBforeignlanguage}[2]{{%
\expandafter\ifx\csname l@#1\endcsname\relax
\typeout{** WARNING: IEEEtran.bst: No hyphenation pattern has been}%
\typeout{** loaded for the language `#1'. Using the pattern for}%
\typeout{** the default language instead.}%
\else
\language=\csname l@#1\endcsname
\fi
#2}}
\providecommand{\BIBdecl}{\relax}
\BIBdecl

\bibitem{Koskinopoulou_waste_sorting}
M.~Koskinopoulou, F.~Raptopoulos, G.~Papadopoulos, N.~Mavrakis, and M.~Maniadakis, ``Robotic waste sorting technology: Toward a vision-based categorization system for the industrial robotic separation of recyclable waste,'' \emph{IEEE Robotics \& Automation Magazine}, vol.~28, no.~2, pp. 50--60, 2021.

\bibitem{zenrobotics_evaluation}
H.~Wilts, B.~R. Garcia, R.~G. Garlito, L.~S. Gómez, and E.~G. Prieto, ``Artificial intelligence in the sorting of municipal waste as an enabler of the circular economy,'' \emph{Resources}, vol.~10, no.~4, 2021.

\bibitem{robot_construction_sorting}
X.~Chen, H.~Huang, Y.~Liu, J.~Li, and M.~Liu, ``Robot for automatic waste sorting on construction sites,'' \emph{Automation in Construction}, vol. 141, p. 104387, 2022.

\bibitem{pick_toss_robot_waste_sorting}
F.~Raptopoulos, M.~Koskinopoulou, and M.~Maniadakis, ``Robotic pick-and-toss facilitates urban waste sorting,'' in \emph{Proc. IEEE International Conference on Automation Science and Engineering (CASE)}, 2020, pp. 1149--1154.

\bibitem{zerowaste}
D.~Bashkirova, M.~Abdelfattah, Z.~Zhu, J.~Akl, F.~Alladkani, P.~Hu, V.~Ablavsky, B.~Calli, S.~A. Bargal, and K.~Saenko, ``{ZeroWaste} dataset: Towards deformable object segmentation in cluttered scenes,'' in \emph{Proc. IEEE/CVF Conference on Computer Vision and Pattern Recognition (CVPR)}, 2022, pp. 21\,115--21\,125.

\bibitem{review_robotic_waste_sorting}
A.~G. Satav, S.~Kubade, C.~Amrutkar, G.~Arya, and A.~Pawar, ``A state-of-the-art review on robotics in waste sorting: scope and challenges,'' \emph{International Journal on Interactive Design and Manufacturing (IJIDeM)}, vol.~17, p. 2789–2806, 2023.

\bibitem{bashkirova2023visda}
D.~Bashkirova, S.~Mishra, D.~Lteif, P.~Teterwak, D.~Kim, F.~Alladkani, J.~Akl, B.~Calli, S.~A. Bargal, K.~Saenko, D.~Kim, M.~Seo, Y.~Jeon, D.-G. Choi, S.~Ettedgui, R.~Giryes, S.~Abu-Hussein, B.~Xie, and S.~Li, ``{VisDA 2022 Challenge}: Domain adaptation for industrial waste sorting,'' in \emph{Proc. Annual Conference on Neural Information Processing Systems (NIPS)}, 2023, pp. 104--118.

\bibitem{rombach2021_diffusion_models}
R.~Rombach, A.~Blattmann, D.~Lorenz, P.~Esser, and B.~Ommer, ``High-resolution image synthesis with latent diffusion models,'' in \emph{Proc. IEEE/CVF Conference on Computer Vision and Pattern Recognition (CVPR)}, 2022, pp. 10\,684--10\,695.

\bibitem{li_eid-gan}
W.~Li, J.~Chen, J.~Cao, C.~Ma, J.~Wang, X.~Cui, and P.~Chen, ``{EID-GAN}: Generative adversarial nets for extremely imbalanced data augmentation,'' \emph{IEEE Transactions on Industrial Informatics}, vol.~19, no.~3, pp. 3208--3218, 2023.

\bibitem{Du_gan_surface}
Z.~Du, L.~Gao, and X.~Li, ``A new contrastive {GAN} with data augmentation for surface defect recognition under limited data,'' \emph{IEEE Transactions on Instrumentation and Measurement}, vol.~72, pp. 1--13, 2023.

\bibitem{ma_GAN-MVA}
P.~Ma, H.~Lu, B.~Yang, and W.~Ran, ``{GAN-MVAE}: A discriminative latent feature generation framework for generalized zero-shot learning,'' \emph{Pattern Recognition Letters}, vol. 155, pp. 77--83, 2022.

\bibitem{xian_F-VAEGAN-D2}
Y.~Xian, S.~Sharma, B.~Schiele, and Z.~Akata, ``{F-VAEGAN-D2}: A feature generating framework for any-shot learning,'' in \emph{Proc. IEEE/CVF Conference on Computer Vision and Pattern Recognition (CVPR)}, 2019, pp. 10\,267--10\,276.

\bibitem{9008794}
L.~Liu, M.~Muelly, J.~Deng, T.~Pfister, and L.-J. Li, ``Generative modeling for small-data object detection,'' in \emph{Proc. IEEE/CVF International Conference on Computer Vision (ICCV)}, 2019, pp. 6072--6080.

\bibitem{9191127}
M.~Hammami, D.~Friboulet, and R.~Kechichian, ``Cycle {GAN}-based data augmentation for multi-organ detection in {CT} images via {YOLO},'' in \emph{Proc. IEEE International Conference on Image Processing (ICIP)}, 2020, pp. 390--393.

\bibitem{li_gan_sem_seg}
D.~Li, J.~Yang, K.~Kreis, A.~Torralba, and S.~Fidler, ``Semantic segmentation with generative models: Semi-supervised learning and strong out-of-domain generalization,'' in \emph{Proc. IEEE/CVF Conference on Computer Vision and Pattern Recognition (CVPR)}, 2021, pp. 8296--8307.

\bibitem{suction_waste_robot}
S.~Um, K.-S. Kim, and S.~Kim, ``Suction point selection algorithm based on point cloud for plastic waste sorting,'' in \emph{Proc. IEEE International Conference on Automation Science and Engineering (CASE)}, 2021, pp. 60--65.

\bibitem{robotic_waste_sorting_system}
M.~Koskinopoulou, F.~Raptopoulos, G.~Papadopoulos, N.~Mavrakis, and M.~Maniadakis, ``Robotic waste sorting technology: Toward a vision-based categorization system for the industrial robotic separation of recyclable waste,'' \emph{IEEE Robotics \& Automation Magazine}, vol.~28, no.~2, pp. 50--60, 2021.

\bibitem{robotic_sorter_kiyokawa}
T.~Kiyokawa, H.~Katayama, Y.~Tatsuta, J.~Takamatsu, and T.~Ogasawara, ``Robotic waste sorter with agile manipulation and quickly trainable detector,'' \emph{IEEE Access}, vol.~9, pp. 124\,616--124\,631, 2021.

\bibitem{construction_demolotion_waste}
Y.~Ku, J.~Yang, H.~Fang, W.~Xiao, and J.~Zhuang, ``Deep learning of grasping detection for a robot used in sorting construction and demolition waste,'' \emph{Journal of Material Cycles and Waste Management}, vol.~23, pp. 84--95, 2021.

\bibitem{grasp_rectangle}
I.~Lenz, H.~Lee, and A.~Saxena, ``Deep learning for detecting robotic grasps,'' \emph{The International Journal of Robotics Research (IJRR)}, vol.~34, no. 4-5, pp. 705--724, 2015.

\bibitem{gan_bengio}
I.~Goodfellow, J.~Pouget-Abadie, M.~Mirza, B.~Xu, D.~Warde-Farley, S.~Ozair, A.~Courville, and Y.~Bengio, ``Generative adversarial networks,'' \emph{Communications of the ACM}, vol.~63, no.~11, p. 139–144, 2020.

\bibitem{mirza2014_cgan}
M.~Mirza and S.~Osindero, ``Conditional generative adversarial nets,'' \emph{arXiv:1411.1784}, 2014.

\bibitem{wgan-arjovsky17a}
M.~Arjovsky, S.~Chintala, and L.~Bottou, ``{W}asserstein generative adversarial networks,'' in \emph{Proc. International Conference on Machine Learning (ICML)}, 2017, pp. 214--223.

\bibitem{karras2018_pggan}
T.~Karras, T.~Aila, S.~Laine, and J.~Lehtinen, ``Progressive growing of gans for improved quality, stability, and variation,'' in \emph{Proc. International Conference on Learning Representations (ICLR)}, 2018.

\bibitem{style_gan}
T.~Karras, S.~Laine, and T.~Aila, ``A style-based generator architecture for generative adversarial networks,'' \emph{IEEE Transactions on Pattern Analysis and Machine Intelligence}, vol.~43, no.~12, pp. 4217--4228, 2021.

\bibitem{stylegan2}
T.~Karras, S.~Laine, M.~Aittala, J.~Hellsten, J.~Lehtinen, and T.~Aila, ``Analyzing and improving the image quality of stylegan,'' in \emph{Proc. IEEE/CVF Conference on Computer Vision and Pattern Recognition (CVPR)}, 2020, pp. 8107--8116.

\bibitem{Bissoto_2021_CVPR}
A.~Bissoto, E.~Valle, and S.~Avila, ``{GAN}-based data augmentation and anonymization for skin-lesion analysis: A critical review,'' in \emph{Proc. IEEE/CVF Conference on Computer Vision and Pattern Recognition (CVPR) Workshops}, 2021, pp. 1847--1856.

\bibitem{covid_Waheed}
A.~Waheed, M.~Goyal, D.~Gupta, A.~Khanna, F.~Al-Turjman, and P.~R. Pinheiro, ``{CovidGAN}: Data augmentation using auxiliary classifier {GAN} for improved {Covid-19} detection,'' \emph{IEEE Access}, vol.~8, pp. 91\,916--91\,923, 2020.

\bibitem{fruit_BIRD}
J.~J. Bird, C.~M. Barnes, L.~J. Manso, A.~Ekárt, and D.~R. Faria, ``Fruit quality and defect image classification with conditional {GAN} data augmentation,'' \emph{Scientia Horticulturae}, vol. 293, p. 110684, 2022.

\bibitem{crop_Fawakherji}
M.~Fawakherji, C.~Potena, A.~Pretto, D.~D. Bloisi, and D.~Nardi, ``Multi-spectral image synthesis for crop/weed segmentation in precision farming,'' \emph{Robotics and Autonomous Systems}, vol. 146, p. 103861, 2021.

\bibitem{xia_gan_inversion}
W.~Xia, Y.~Zhang, Y.~Yang, J.-H. Xue, B.~Zhou, and M.-H. Yang, ``{GAN} inversion: A survey,'' \emph{IEEE Transactions on Pattern Analysis and Machine Intelligence}, vol.~45, no.~3, pp. 3121--3138, 2023.

\bibitem{datasetgan_zhang}
Y.~Zhang, H.~Ling, J.~Gao, K.~Yin, J.-F. Lafleche, A.~Barriuso, A.~Torralba, and S.~Fidler, ``{DatasetGAN}: Efficient labeled data factory with minimal human effort,'' in \emph{Proc. IEEE/CVF Conference on Computer Vision and Pattern Recognition (CVPR)}, 2021, pp. 10\,145--10\,155.

\bibitem{lim2017_geometric_gan}
J.~H. Lim and J.~C. Ye, ``Geometric {GAN},'' \emph{arXiv:1705.02894}, 2017.

\bibitem{chao_contrained_GAN}
X.~Chao, J.~Cao, Y.~Lu, Q.~Dai, and S.~Liang, ``Constrained generative adversarial networks,'' \emph{IEEE Access}, vol.~9, pp. 19\,208--19\,218, 2021.

\bibitem{perceptual_loss}
R.~Zhang, P.~Isola, A.~A. Efros, E.~Shechtman, and O.~Wang, ``The unreasonable effectiveness of deep features as a perceptual metric,'' in \emph{Proc. IEEE/CVF Conference on Computer Vision and Pattern Recognition (CVPR)}, 2018, pp. 586--595.

\bibitem{labml}
\BIBentryALTinterwordspacing
N.~W. Varuna~Jayasiri, ``labml.ai annotated paper implementations,'' 2020. [Online]. Available: \url{https://nn.labml.ai/}
\BIBentrySTDinterwordspacing

\bibitem{gradient_panality}
I.~Gulrajani, F.~Ahmed, M.~Arjovsky, V.~Dumoulin, and A.~Courville, ``Improved training of {Wasserstein GANs},'' in \emph{Proc. Annual Conference on Neural Information Processing Systems (NIPS)}, 2017, p. 5769–5779.

\bibitem{Iakubovskii_smp}
P.~Iakubovskii, ``Segmentation models pytorch,'' \url{https://github.com/qubvel/segmentation_models.pytorch}, 2019.

\bibitem{fpn}
C.~Wang and C.~Zhong, ``Adaptive feature pyramid networks for object detection,'' \emph{IEEE Access}, vol.~9, pp. 107\,024--107\,032, 2021.

\bibitem{deeplabv3plus2018}
L.-C. Chen, Y.~Zhu, G.~Papandreou, F.~Schroff, and H.~Adam, ``Encoder-decoder with atrous separable convolution for semantic image segmentation,'' in \emph{Proc. The European Conference on Computer Vision (ECCV)}, 2018, p. 833–851.

\bibitem{resnet}
K.~He, X.~Zhang, S.~Ren, and J.~Sun, ``Deep residual learning for image recognition,'' in \emph{Proc. IEEE Conference on Computer Vision and Pattern Recognition (CVPR)}, 2016, pp. 770--778.

\bibitem{tan2019efficientnet}
M.~Tan and Q.~Le, ``{EfficientNet}: Rethinking model scaling for convolutional neural networks,'' in \emph{Proc. International Conference on Machine Learning (ICML)}, 2019, pp. 6105--6114.

\bibitem{mobilenetv3}
A.~Howard, M.~Sandler, B.~Chen, W.~Wang, L.~Chen, M.~Tan, G.~Chu, V.~Vasudevan, Y.~Zhu, R.~Pang, H.~Adam, and Q.~Le, ``Searching for {MobileNetV3},'' in \emph{Proc. IEEE/CVF International Conference on Computer Vision (ICCV)}, 2019, pp. 1314--1324.

\bibitem{mobilevitv2}
S.~Mehta and M.~Rastegari, ``Separable self-attention for mobile vision transformers,'' \emph{arXiv:2206.02680}, 2022.

\bibitem{imagenet}
O.~Russakovsky, J.~Deng, H.~Su, J.~Krause, S.~Satheesh, S.~Ma, Z.~Huang, A.~Karpathy, A.~Khosla, M.~Bernstein, A.~C. Berg, and L.~Fei-Fei, ``Imagenet large scale visual recognition challenge,'' \emph{International Journal of Computer Vision (IJCV)}, vol. 115, no.~3, pp. 211--252, 2015.

\end{thebibliography}

\end{document}